%% file: samplepaper.tex
\documentclass[runningheads]{llncs}
\usepackage[T1]{fontenc}
\usepackage{graphicx}
\begin{document}
\title{How to Train Your CheXDragon:\\Training Chest X-Ray Models for Transfer to Novel Tasks and Healthcare Systems}


%


\titlerunning{How to Train Your CheXDragon}
%
\author{
    Cara Van Uden\inst{1,2} \and
    Jeremy Irvin\inst{1,2} \and
    Mars Huang\inst{2,3} \and
    Nathan Dean\inst{4,5} \and
    Jason Carr\inst{5} \and
    Andrew Ng\inst{1} \and
    Curtis Langlotz\inst{2}
}

%
\authorrunning{C. Van Uden et al.}
%

\institute{
    Department of Computer Science, Stanford University, Stanford, CA, USA \and
    Center for Artificial Intelligence in Medicine and Imaging, Stanford University, Palo Alto, CA, USA \and
    Department of Biomedical Data Science, Stanford University, Stanford, CA, USA \and
    Division of Respiratory, Critical Care, and Occupational Pulmonary Medicine, University of Utah, Salt Lake City, UT, USA \and
    Department of Pulmonary and Critical Care Medicine, Intermountain Medical Center, Salt Lake City, UT, USA
}
%
\maketitle              
\input{templates/abstract}
\input{templates/introduction}
\input{templates/related_work}
\input{templates/methods}
\input{templates/experiments}
\input{templates/results_discussion}
\input{templates/conclusion}
\input{templates/acknowledgements}

\pagebreak
%
%
%
\bibliographystyle{splncs04}
\bibliography{bibliography}

\input{templates/appendix}

\end{document}

%% file: templates/abstract.tex
\begin{abstract}

Self-supervised learning (SSL) enables label efficient training for machine learning models. This is essential for domains such as medical imaging, where labels are costly and time-consuming to curate. However, the most effective supervised or SSL strategy for transferring models to different healthcare systems or novel tasks is not well understood. In this work, we systematically experiment with a variety of supervised and self-supervised pretraining strategies using multimodal datasets of medical images (chest X-rays) and text (radiology reports). We then evaluate their performance on data from two external institutions with diverse sets of tasks. In addition, we experiment with different transfer learning strategies to effectively adapt these pretrained models to new tasks and healthcare systems. Our empirical results suggest that multimodal SSL gives substantial gains over unimodal SSL in performance across new healthcare systems and tasks, comparable to models pretrained with full supervision. We demonstrate additional performance gains with models further adapted to the new dataset and task, using multimodal domain-adaptive pretraining (DAPT), linear probing then finetuning (LP-FT), and both methods combined. We offer suggestions for alternative models to use in scenarios where not all of these additions are feasible. Our results provide guidance for improving the generalization of medical image interpretation models to new healthcare systems and novel tasks.

\end{abstract}

%% file: templates/introduction.tex
\section{Introduction}


Medical imaging plays a crucial role in the diagnosis and treatment of various diseases, and the development of accurate and robust models for automated image analysis has the potential to improve the care of millions of patients worldwide. Many prior studies have successfully automated chest X-ray interpretation on a variety of clinical tasks \cite{rajpurkar2017chexnet,rajpurkar2018chexnext,rajpurkar2018chexnext,aggarwal2021diag,esteva2021deep}, often achieving performance that is comparable to or better than clinical experts \cite{rajpurkar2018chexnext,irvin2019chexpert,wehbe2021deepcovid,tiu2022chexzero}. Much of this success can be attributed to the use of large, labeled datasets of images \cite{johnson2019mimiccxr,irvin2019chexpert,bustos2020padchest,feng2021candidptx}. However, transferring a model trained with data from a specific healthcare system and task set to new clinical settings 
often results in a drop in performance due to \textit{distribution shifts} \cite{karani2022thesis}.
This phenomenon is a limiting factor for the widespread transfer and adoption of these medical imaging models. 

\textit{Self-supervised learning} (SSL)
has been demonstrated to be an effective way to combat distribution shifts for both natural \cite{chen2020simclrv2,radford2021clip,fang2022data,nguyen2022quality} and medical \cite{sowrirajan2021mococxr,vu2021medaug,zhang2020contrastive,huang2021gloria,tiu2022chexzero} images. While prior studies \cite{azizi2021bigmic,ke2021chextransfer,azizi2022robust,chowdhury2021applying,shurrab2022self} have investigated the impact of different strategies for pretraining, SSL or transfer learning for medical imaging tasks, these studies have not compared the impact of more recent advances in model pretraining (i.e. multimodal SSL) or the intersection of different pretraining and transfer strategies. Systematic evaluations and benchmarking of these methods are necessary for guiding the practical implementation and deployment of medical imaging models, especially for new tasks and healthcare systems.

In this work, we conduct a thorough set of experiments to address this gap. Our experiments aim to answer the following questions about transferring medical imaging models to novel medical tasks and new hospital systems, with a focus on chest X-ray interpretation:
\begin{enumerate}
    \item How do different pretraining strategies affect downstream performance? We find that multimodal SSL substantially improves downstream performance over unimodal SSL, comparable to performance with full supervision.
    \item What effect do the size and characteristics of the pretraining datasets have on downstream performance? Our results suggest that using both the pretraining and downstream datasets during pretraining (multimodal "domain-adaptive pretraining", or DAPT) improves downstream performance.
    \item Which real-world pretraining and finetuning scenarios provide the highest downstream performance? We demonstrate that the best performance is achieved with a model that is heavily adapted to the new dataset and task, using both multimodal domain-adaptive self-supervised pretraining (multimodal DAPT) and linear probing, then finetuning (LP-FT).
\end{enumerate}

We propose a "decision tree" (Figure \ref{fig:decision_tree}) for a recommended chest X-ray model pretraining and transfer learning workflow. Given our experimentation on multiple downstream datasets, we hope that this workflow is applicable for many chest X-ray, and potentially other medical imaging, datasets and tasks.

%% file: templates/methods.tex
\section{Methods}

\subsection{Datasets}


\begin{table}[]
\centering
\resizebox{0.80\textwidth}{!}{
\begin{tabular}{l|cccc}
\hline
\textbf{Dataset}    & \textbf{ \texttt{\#}Train } & \textbf{\texttt{\#}Val } & \textbf{\texttt{\#}Test } & \textbf{\texttt{\#}Total} \\ \hline
\textit{CheXpert}   & 218,414 & 5,000 & 1,000  & 224,414 \\ \hline
Atelectasis          & 65,634  & 1,481 & 200    & 67,315   \\
Cardiomegaly         & 34,290  & 797   & 200    & 35,287   \\
Consolidation        & 41,546  & 979   & 200    & 42,725   \\
Edema      & 63,798  & 1,432 & 200    & 65,430   \\
Pleural Effusion     & 95,620  & 2,195 & 200    & 98,015   \\ \hline
\textit{Intermountain}        & 19,028  & 3,361 & 2,256  & 21,184  \\ \hline
Pneumonia  & 12,499  & 736   & 630    & 13,865   \\
Pneumonia+Pleural Effusion     & 1,628   & 95    & 61     & 1,784    \\
Pneumonia+Multifocal Pneumonia & 4,390   & 376   & 419    & 5,185    \\ \hline
\textit{CANDID-PTX}           & 15,387  & 1,923 & 1,923  & 19,237  \\ \hline
Pneumothorax         & 2,553   & 323   & 320    & 3,196    \\
Rib Fracture         & 254     & 41    & 40     & 335     \\
Chest Tube           & 1,149   & 140   & 134    & 1,423    \\ \hline
\end{tabular}
}
\caption{Number of studies and tasks in the three chest X-ray datasets used for experimentation. For each dataset, we report the total number of studies per split as well as the number of positive studies for each task per split. For CheXpert, unless stated otherwise, we use all 14 of the original tasks \cite{irvin2019chexpert} during pretraining, and the subset of 5 tasks described above (the CheXpert competition tasks) during testing.}
\label{table:data_stats}
\end{table}

\begin{figure}[h!]
    \centering
    \includegraphics[width = 1\textwidth]{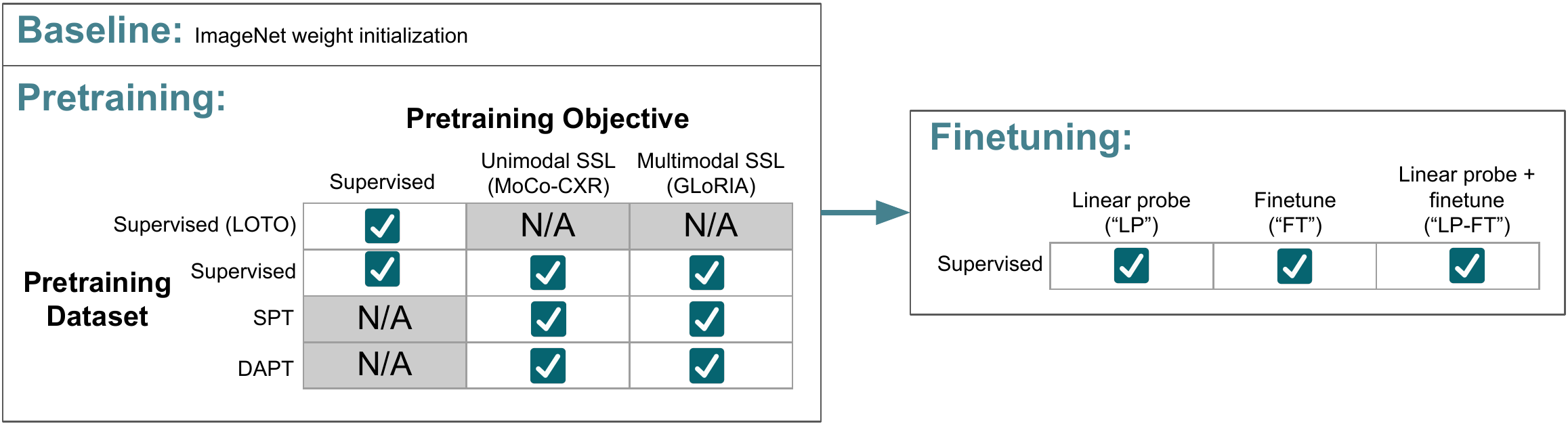}
    \caption{Summary of experiments. Appendix describes invalid experiments ("N/A").}
    \label{fig:experiments}
\end{figure}

\subsubsection{CheXpert}
We use the CheXpert \cite{irvin2019chexpert} dataset as the primary pretraining dataset. CheXpert is a large public dataset of 224,316 chest X-ray studies from 65,240 patients, collected from inpatients and outpatients at Stanford Health Care between 2002 and 2017. Each study has an associated radiology report 
dictated by a radiologist during clinical care. Each study is also labeled for 14 binary clinical observations with an automatic rule-based labeler \cite{irvin2019chexpert}, which we use as our pretraining tasks. We use the 5 CheXpert competition
tasks, identified by a consensus of radiologists, as our test tasks. These tasks are Atelectasis, Cardiomegaly, Consolidation, Edema, and Pleural Effusion. For CheXpert and all following datasets, we report the tasks and number of studies in Table~\ref{table:data_stats}.

\subsubsection{Intermountain}
We use another dataset from Intermountain Healthcare \cite{irvin2022chexed} as one of the external healthcare system ("downstream") datasets. The dataset contains 21,184 chest X-ray studies collected at Intermountain Health (Utah, USA) outpatient centers (including 5 emergency departments and 22 urgent care centers) between 2009 and 2021. Like CheXpert, each study has an associated radiology report dictated by a radiologist for clinical care. A physician labeled each study for 3 pneumonia-related clinical observations: Pneumonia, Pneumonia+Pleural Effusion, and Pneumonia+Multifocal Pneumonia.

\subsubsection{CANDID-PTX}
We use CANDID-PTX \cite{feng2021candidptx} as the second downstream dataset. CANDID-PTX is a publicly available dataset containing 19,237 chest X-ray studies, collected at Dunedin Hospital (Dunedin, NZ) between 2010 and 2020. Each study has an associated radiology report dictated by a radiologist for clinical care. A physician or consensus of physicians labeled each study for 3 clinical observations: Pneumothorax, Acute Rib Fracture, and Intercostal Chest Tube.

\begin{table}
\centering
\resizebox{0.96\textwidth}{!}{
\begin{tabular}{|c|c c c c|c c c c c c|}
\hline
 \textbf{\shortstack{Labeled\\Fraction}} & 
 \textbf{\shortstack{Pretraining\\Dataset}} & \textbf{\shortstack{Pretraining\\Objective}} & 
 \textbf{\shortstack{NO-DAPT/\\DAPT?}} &
 \textbf{\shortstack{LP/\\LP-FT?}} &
 \textbf{\shortstack{Macro-\\AUROC}} &  \textbf{\shortstack{Micro-\\AUROC}} & \textbf{\shortstack{Macro-\\AUPRC}} & \textbf{\shortstack{Micro-\\AUPRC}} & \textbf{\shortstack{Macro-\\F1}} &  \textbf{\shortstack{Micro-\\F1}} \\
\hline
\multirow{12}{*}{0.01}  
&  ImageNet & Sup & NO-DAPT & LP & 0.548 & 0.721 & 0.152 & 0.234 & 0.073 & 0.187 \\
&  ImageNet & Sup & DAPT & LP & 0.705 & 0.781 &  0.34 & 0.462 & 0.206 & 0.338 \\
&  ImageNet & Sup & NO-DAPT & LP-FT & 0.587 & 0.536 & 0.177 & 0.182 &0.2 & 0.246 \\
&  ImageNet & Sup & DAPT & LP-FT &  0.86 & 0.906 & 0.489 & 0.651 & 0.388 & 0.594 \\
\cdashline{2-11}
&  CheXpert & Sup & NO-DAPT & LP & 0.812 &  0.85 & 0.423 & 0.567 & 0.168 &  0.386 \\
&  CheXpert & Sup & DAPT & LP &  0.77 & 0.855 & 0.456 & 0.601 & 0.171 & 0.375 \\
&  CheXpert & Sup & NO-DAPT & LP-FT &  \underline{0.87} & 0.896 & 0.503 & 0.647 & 0.315 &  0.562 \\
&  CheXpert & Sup & DAPT & LP-FT & 0.866 & \underline{0.911} &  \underline{0.55} & \underline{0.701} & \underline{0.396} & \underline{0.632} \\
\cdashline{2-11}
&  CheXpert & GLoRIA & NO-DAPT & LP & 0.672 & 0.804 & 0.327 & 0.477 & 0.096 & 0.234 \\
&  CheXpert & GLoRIA & DAPT & LP & 0.817 & 0.888 & 0.492 & 0.656 & 0.232 & 0.476 \\
&  CheXpert & GLoRIA & NO-DAPT & LP-FT & 0.837 & 0.876 & 0.461 & 0.601 & 0.278 & 0.513 \\
&  CheXpert & GLoRIA & DAPT & LP-FT & \textbf{0.874} &  \textbf{0.92*} & \textbf{0.555*} & \textbf{0.718*} & \textbf{0.416} & \textbf{0.649*} \\
\hline
\multirow{12}{*}{0.1}  
&  ImageNet & Sup & NO-DAPT & LP & 0.709 & 0.795 & 0.273 & 0.352 & 0.102 & 0.228 \\
&  ImageNet & Sup & DAPT & LP &0.9 & 0.931 & 0.556 & 0.725 & 0.471 & 0.663 \\
&  ImageNet & Sup & NO-DAPT & LP-FT & 0.821 & 0.854 & 0.417 & 0.545 & 0.377 & 0.538 \\
&  ImageNet & Sup & DAPT & LP-FT & 0.901 & 0.934 & 0.575 & 0.735 & 0.519 & 0.679 \\
\cdashline{2-11}
&  CheXpert & Sup & NO-DAPT & LP & 0.907 & 0.925 &  0.59 & 0.721 & 0.491 & 0.653 \\
&  CheXpert & Sup & DAPT & LP & 0.921 & 0.945 & 0.635 & 0.787 & 0.556 & 0.718 \\
&  CheXpert & Sup & NO-DAPT & LP-FT & 0.919 & 0.938 & 0.638 & 0.763 & 0.567 & 0.704 \\
&  CheXpert & Sup & DAPT & LP-FT & 0.925 & 0.947 & \underline{0.648} & \underline{0.796} & \underline{0.575} & \underline{0.727} \\
\cdashline{2-11}
&  CheXpert & GLoRIA & NO-DAPT & LP & 0.9 & 0.922 & 0.579 & 0.713 & 0.469 & 0.641 \\
&CheXpert & GLoRIA & DAPT & LP & \underline{0.926} & \underline{0.948} & 0.629 &  0.79 & 0.555 &  0.72 \\
&  CheXpert & GLoRIA & NO-DAPT & LP-FT & 0.908 & 0.935 &  0.62 &  0.76 & 0.545 & 0.697 \\
&  CheXpert & GLoRIA & DAPT & LP-FT &  \textbf{0.93} & \textbf{0.949} & \textbf{0.651} & \textbf{0.798} & \textbf{0.577} & \textbf{0.728} \\
\hline
\multirow{12}{*}{1}  
&  ImageNet & Sup & NO-DAPT & LP & 0.815 & 0.852 & 0.398 &  0.47 & 0.275 & 0.419 \\
&  ImageNet & Sup & DAPT & LP & 0.914 & 0.938 & 0.597 & 0.746 & 0.513 & 0.682 \\
&  ImageNet & Sup & NO-DAPT & LP-FT & 0.908 & 0.927 & 0.609 & 0.735 & 0.546 & 0.693 \\
&  ImageNet & Sup & DAPT & LP-FT & 0.923 & 0.942 & 0.651 & 0.773 & 0.617 & 0.719 \\
\cdashline{2-11}
&  CheXpert & Sup & NO-DAPT & LP & 0.926 & 0.939 & 0.649 & 0.764 & 0.554 & 0.689 \\
&  CheXpert & Sup & DAPT & LP & 0.936 & \underline{0.952} & 0.665 & 0.806 & 0.586 & 0.731 \\
&  CheXpert & Sup & NO-DAPT & LP-FT & 0.937 &  0.95 & \textbf{0.703*} & 0.803 & 0.655 & 0.748 \\
&  CheXpert & Sup & DAPT & LP-FT & 0.936 & \underline{0.952} & \underline{0.7} & \underline{0.816} & \textbf{0.662} & \underline{0.758} \\
\cdashline{2-11}
&  CheXpert & GLoRIA & NO-DAPT & LP & 0.921 & 0.934 & 0.627 & 0.746 & 0.545 & 0.677 \\
&  CheXpert & GLoRIA & DAPT & LP & \underline{0.938} & \underline{0.952} &  0.66 & 0.805 & 0.585 & 0.728 \\
&  CheXpert & GLoRIA & NO-DAPT & LP-FT & 0.932 & 0.945 & 0.661 & 0.775 & 0.627 & 0.732 \\
&  CheXpert & GLoRIA & DAPT & LP-FT & \textbf{0.946} & \textbf{0.955} & \textbf{0.703*} & \textbf{0.822} & \underline{0.659} & \textbf{0.759} \\
\hline
\end{tabular}

}
 \caption{
     Results for ImageNet-Supervised, CheXpert-Supervised, and CheXpert-GLoRIA models at label fractions of 0.01, 0.1, and 1. For each initial pretrained model type, we show results with a simple linear probe ("+LP"); with linear probing, then finetuning ("+LP-FT"); with multimodal domain-adaptive pretraining and a linear probe ("+DAPT+LP"); and with multimodal domain-adaptive pretraining and linear probing, then finetuning ("+DAPT+LP-FT"). The best result for each label fraction is \textbf{bolded} and next-best is \underline{underlined}. There are no statistically significant differences (paired samples t-test with Bonferroni correction, 95\% CI) between the best model result over the next-best model result, so we denote a statistically significant difference between the best model result and the third-best with "*".
 }
 \label{table:results}
\end{table}

\subsection{Model pretraining and transfer learning}
Our aim is to rigorously investigate the effect of pretraining and transfer strategies on the performance of medical imaging interpretation models in new clinical settings and on novel tasks. To do this, we first pretrain each model using one combination of pretraining dataset and pretraining objective. We then keep the image encoder from that pretraining step, and \textit{transfer} to the downstream dataset. We use a DenseNet-121 \cite{huang2017densenet} for the image encoder. This has demonstrated success in several previous studies \cite{rajpurkar2018chexnext,irvin2019chexpert,irvin2022chexed}. For multimodal SSL, we use BioClinicalBERT \cite{alsentzer2019bioclinical} as the text encoder. Figure \ref{fig:experiments} contains a summary of our experiments. Please see the Appendix for all implementation details.

\subsubsection{Pretraining datasets}
\label{varying_pretraining_datasets}
We compare three methods for varying the pretraining dataset: \textit{vanilla pretraining}, where we pretrain on the large pretraining dataset; \textit{self-pretraining} \cite{krishna2022spt} (SPT), where we pretrain on the smaller downstream dataset; \textit{domain-adaptive pretraining} \cite{gururangan2022dapt} (DAPT), where we first pretrain on CheXpert, then pretrain on the smaller downstream dataset.

\subsubsection{Pretraining objectives}
\label{varying_pretraining_obj}
We compare three pretraining objectives: fully-supervised pretraining, unimodal (image-only) self-supervised pretraining with MoCo-CXR \cite{sowrirajan2021mococxr}, and multimodal (image+text) self-supervised pretraining with GLoRIA \cite{huang2021gloria}. Fully-supervised training uses chest X-rays and clinical observation labels, unimodal self-supervised pretraining uses only contrastive learning of chest X-rays, and multimodal self-supervised pretraining contrasts both chest X-rays and the impression section of their accompanying radiology report. For fully-supervised pretraining, we perform "leave-one-task-out" (LOTO) ablations to investigate how "task novelty", or a downstream task appearing/not appearing in the pretraining task set, affects downstream performance on that task.

\subsubsection{Supervised transfer}
\label{sup_trans}
In the transfer step, we take the pretrained image encoder from each of the above pretraining methods and train either only a linear classifier on top of the encoder ("linear probing" or "LP") or first a linear classifier and \textit{then} the entire model end-to-end ("linear probe $\rightarrow$ finetuning", or "LP-FT") \cite{kumar2022finetuning}. We train with 1\%, 5\%, 10\%, 20\%. or 100\% (the "label fraction") of the available labeled downstream dataset to examine the impact of dataset size on performance. For every label fraction, we train 5 models with different random seeds to create 95\% confidence intervals for each experimental result.

%% file: templates/experiments.tex


%% file: templates/results_discussion.tex
\section{Results and Discussion}
\begin{figure}[h!]
    \centering
    \includegraphics[width = 0.9\textwidth]{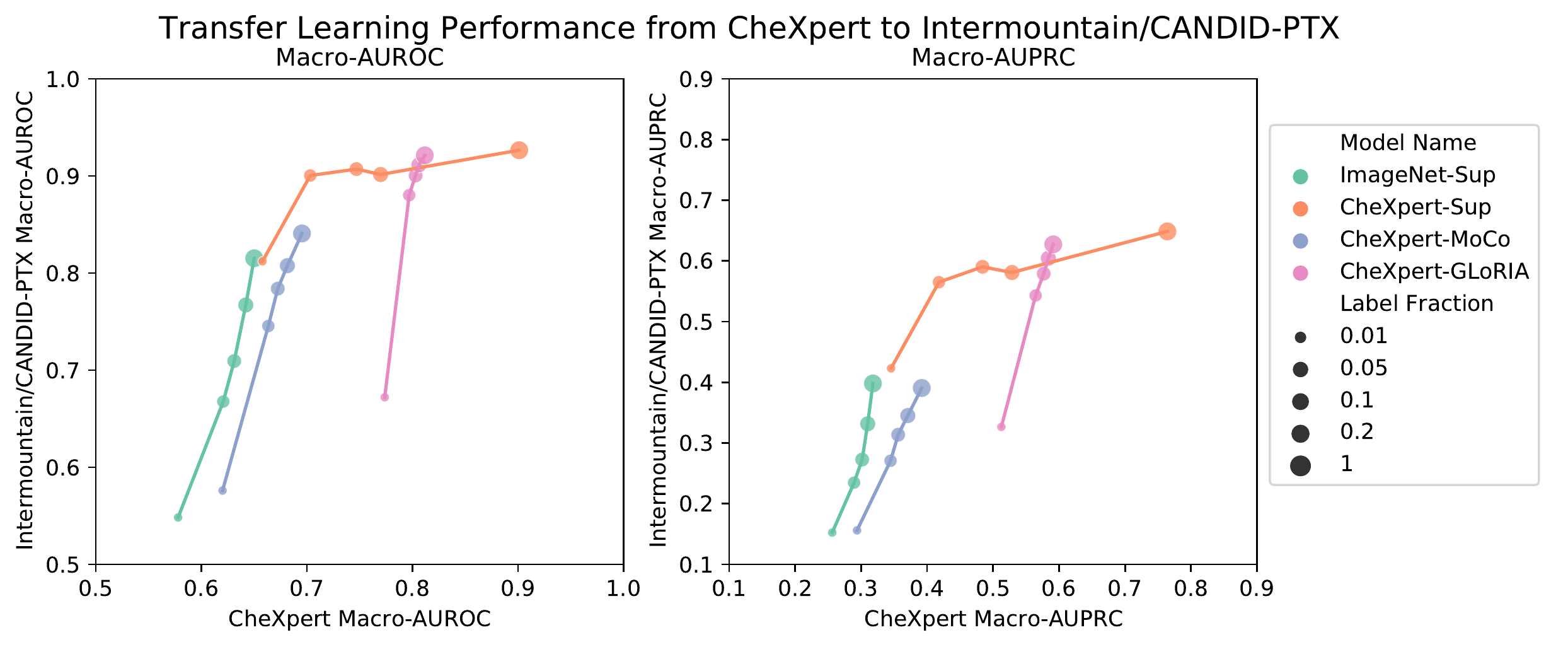}
    \caption{Performance of supervised and self-supervised (uni-, multimodal) pretrained models before (on CheXpert) and after (on Intermountain/CANDID-PTX) transfer learning via linear probing, measured via macro-AUROC (L) and macro-AUPRC (R).}
    \label{fig:transfer_learning}
\end{figure}

\begin{figure}[h!]
    \centering
    \includegraphics[width = 0.9\textwidth]{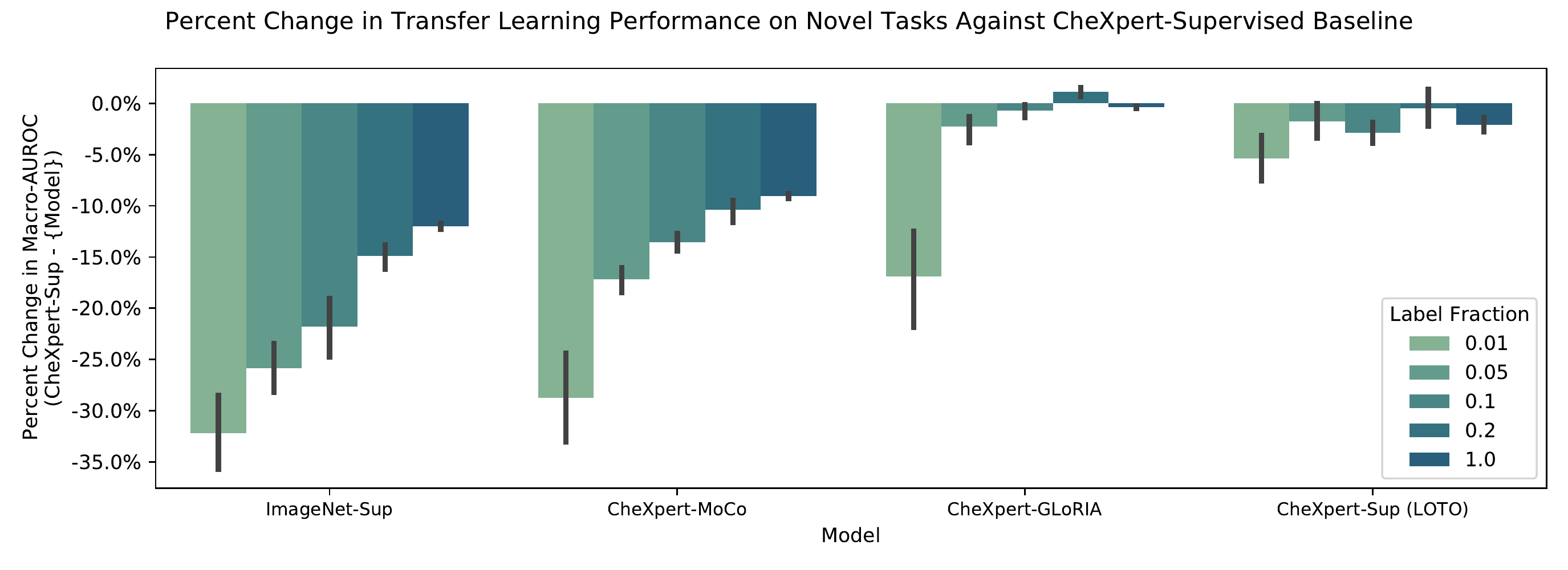}
    \caption{Performance differences between CheXpert-Sup and ImageNet-Sup, LOTO-CheXpert-Sup, CheXpert-MoCo, and CheXpert-GloRIA. Error bars denote 95\% CI over 5 runs.}
    \label{fig:loto_barplot}
\end{figure}

\begin{figure}[h!]
    \centering
    \includegraphics[width = 0.9\textwidth]{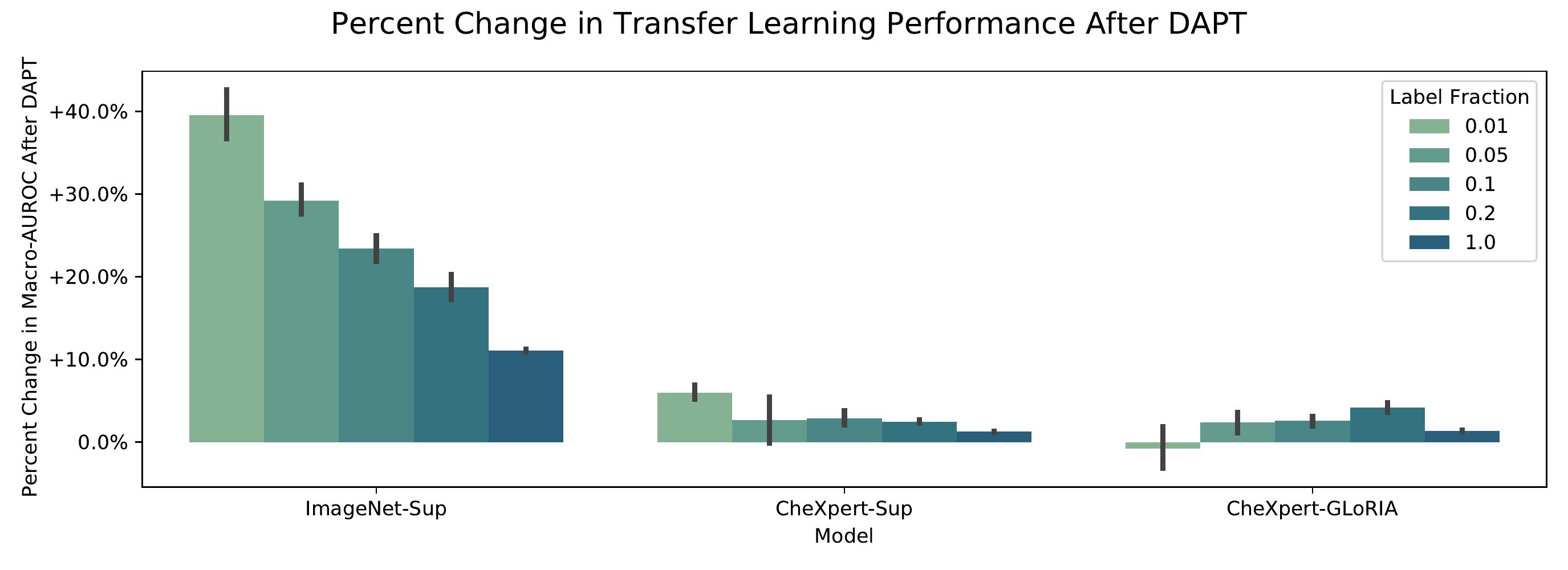}
    \caption{DAPT impact on downstream performance. Error bars over 5 runs (95\% CI).}
    \label{fig:dapt_percent_increase_combine_datasets}
\end{figure}

\begin{figure}[h!]
    \centering
    \includegraphics[width = 0.95\textwidth]{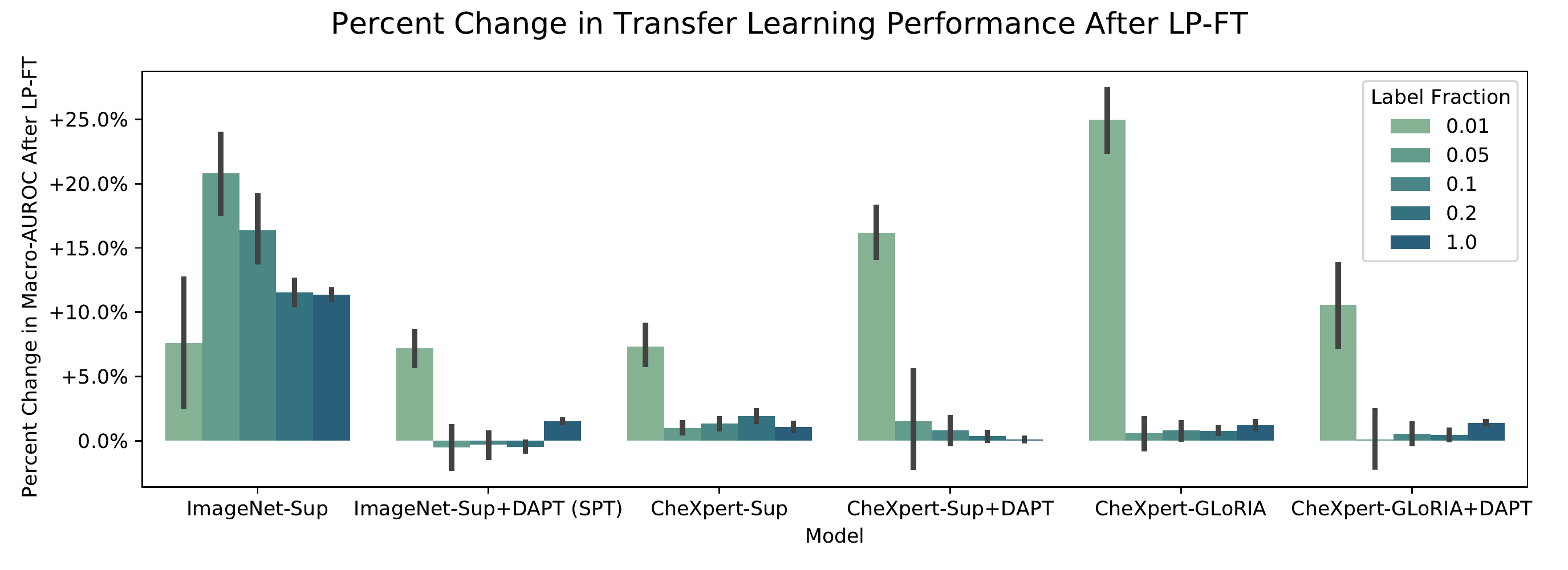}
    \caption{LP-FT impact on downstream performance. Error bars over 5 runs (95\% CI).}
    \label{fig:end_to_end_percent_increase_combine_datasets}
\end{figure}

\subsection{Performance with no further pretraining (NO-DAPT+LP)}
\subsubsection{Performance on novel datasets}
\label{transfer_section}

First, we focus on performance of self-supervised and supervised-pretrained models in a simple setting: first pretraining on a large dataset (CheXpert) and then linear probing on a downstream dataset (Intermountain or CANDID-PTX). We compare a model's performance on the pretraining dataset with its performance on the downstream dataset. As seen in Figure \ref{fig:transfer_learning}, across all label fractions, ImageNet-Supervised and CheXpert-MoCo (unimodal SSL) models underperform against CheXpert-GLoRIA (multimodal SSL) models. However, compared to models pretrained with supervision on CheXpert, CheXpert-GLoRIA models underperform for smaller label fractions (0.01, 0.05, 0.1) and match performance for larger label fractions (0.2, 1).


\subsubsection{Performance on novel tasks}
\label{loto_section}

We investigate the impact of task novelty on downstream task performance. "Novel" tasks to a supervised-pretrained model are downstream tasks that did not appear in the supervised pretraining set. We pretrained supervised models on CheXpert and held out one or two tasks for each model ("LOTO-CheXpert-Sup"). We pair each of these models with their corresponding downstream task (details in Appendix). As seen in Figure \ref{fig:loto_barplot}, LOTO-CheXpert-Sup models outperform CheXpert-GLoRIA (multimodal SSL) on novel downstream tasks at small label fractions (0.01, 0.05). For larger label fractions, CheXpert-GLoRIA models outperform LOTO-CheXpert-Sup models.

\subsection{Performance after further pretraining on the downstream dataset (DAPT+LP)}
\label{dapt_section}

Next, we focus on how further pretraining on the downstream dataset affects downstream task performance. As seen in Table \ref{table:results} and Figure \ref{fig:dapt_percent_increase_combine_datasets}, after linear probing, multimodal domain-adaptive pretrained (DAPT) models outperform their non-DAPT counterparts. For almost all label fractions and all pretrained model types, DAPT substantially improves downstream task performance. 


\subsection{Performance after further pretraining on the downstream dataset and linear probing+finetuning (DAPT+LP-FT)}
\label{dapt_lpft_section}

We investigate how heavily adapting to the downstream dataset (via both DAPT and LP-FT) affects downstream task performance by adding end-to-end finetuning to all linear probe (LP) models described in \ref{dapt_section}. As seen in Table \ref{table:results} and Figure \ref{fig:end_to_end_percent_increase_combine_datasets}, LP-FT matches or outperforms LP across all explored pretrained models. In particular, LP-FT is label-efficient and gives substantial performance improvements at the lowest investigated label fraction (0.01). As a note, we found that randomly initializing a linear probe and subsequently end-to-end finetuning (equivalent to "FT" in Figure \ref{fig:experiments}) \textit{degrades} performance for almost all models and label fractions compared to a standard linear probe; see details in the Appendix.

\begin{figure}[h!]
    \centering
    \includegraphics[width = 0.90\textwidth]{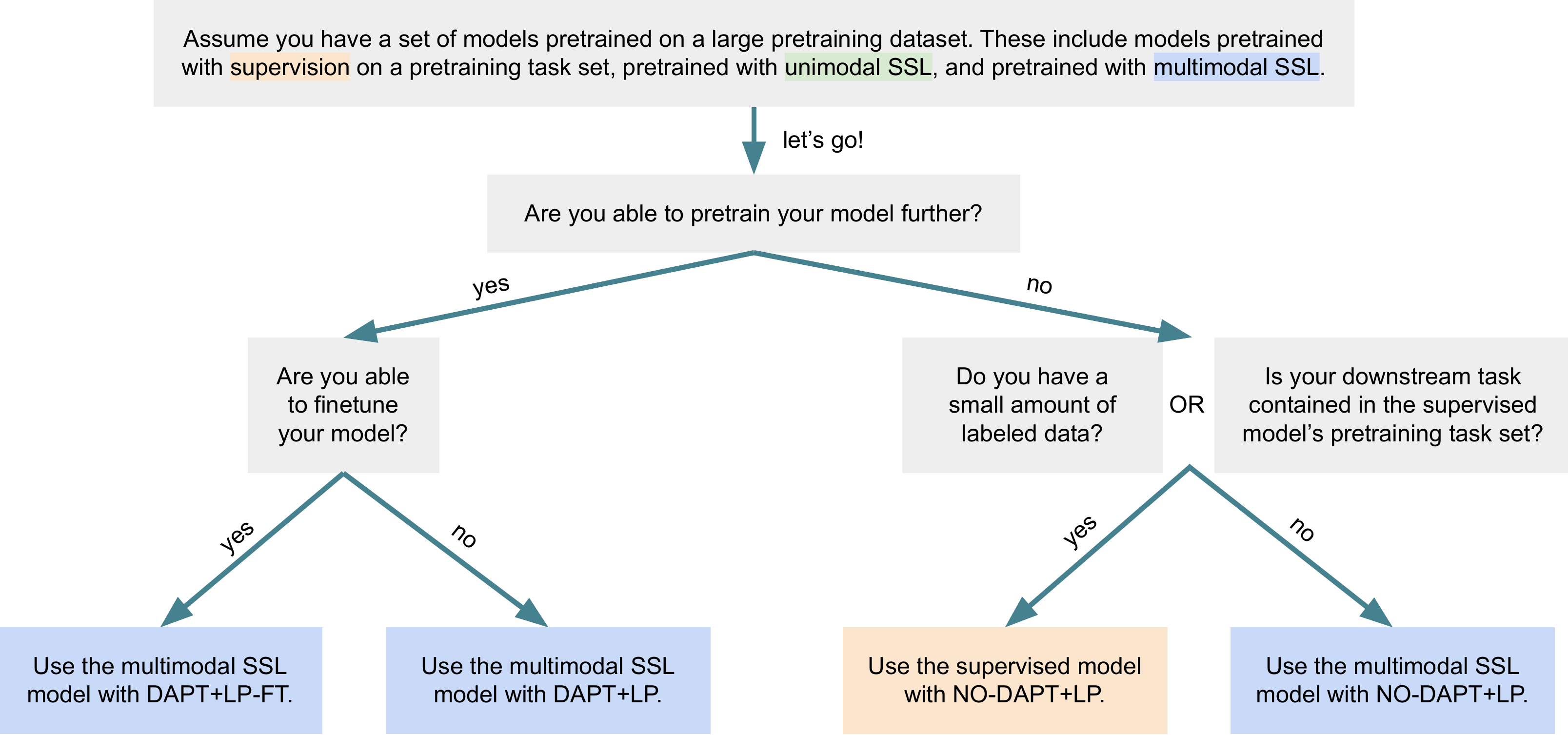}
    \caption{Decision tree for a pretraining and transfer learning workflow. We recommend starting with either a fully-supervised pretrained model or a multimodal SSL-pretrained model. We find that a multimodal SSL model with DAPT+LP-FT achieves the best performance for all label fractions, downstream datasets, and downstream tasks.}
    \label{fig:decision_tree}
\end{figure}

\subsection{"Decision tree" for a pretraining and transfer learning workflow}

We demonstrate that the best performance is achieved with a model that is heavily adapted to the new dataset and task, using both multimodal domain-adaptive pretraining and LP-FT finetuning. In Figure \ref{fig:decision_tree}, we also offer a "decision tree" of suggestions for alternative modeling choices to make in other scenarios. 

%% file: templates/conclusion.tex
\section{Conclusion}

In this work, we rigorously investigate chest X-ray model pretraining and transfer learning strategies for new clinical settings and novel tasks. Given a set of pretrained models, we investigate a set of real-world scenarios featuring novel datasets and tasks in which we might use these pretrained models: with no further pretraining and linear probing (NO-DAPT+LP), with further pretraining and linear probing (DAPT+LP), and with both further pretraining and linear probing, then end-to-end finetuning (DAPT+LP-FT). We find that multimodal SSL pretraining outperforms unimodal SSL pretraining methods and matches supervised pretraining when transferring to data from new healthcare systems and novel tasks. Finally, we find that domain-specific pretraining (via DAPT) and finetuning (via LP-FT), both separately and combined, substantially improve downstream task performance. Our work provides tangible guidance for adapting chest X-ray interpretation, and potentially other medical imaging, models to new clinical settings and diagnostic tasks. We hope that these findings support the development of medical imaging models that can improve patient care.

%% file: templates/acknowledgements.tex
\vspace{12.5cm}
\section{Acknowledgements}
This work was supported in part by the Agency for Health Research and Quality under grant \#5R18HS02688. The authors acknowledge the contributions of Jeremy Irvin, Pranav Rajpurkar, and Stanford Health Care for the creation of the CheXpert dataset, Intermountain Healthcare for the creation of the Intermountain dataset, and SiJing Feng and Dunedin Hospital for the creation of the CANDID-PTX dataset. 

%% file: templates/appendix.tex
\section{Appendix}





\begin{table}[h!]
\centering
\resizebox{1\textwidth}{!}{
\begin{tabular}{l | l}
    \hline
    \textbf{\shortstack{Invalid Experiment Name}} & \textbf{Why?} \\
    \hline
    "LOTO" SSL PT & SSL pretraining has no tasks to "leave out" \\
    Supervised PT on downstream dataset & equivalent to the supervised transfer step \\
    \hline
\end{tabular}
}
\caption{
    Description of invalid experiments from Figure \ref{fig:experiments}. "Leave-one-task-out" ("LOTO") SSL pretraining is invalid because there are no tasks to "leave out" in SSL pretraining. Supervised pretraining on the downstream dataset (via SPT, DAPT) is invalid because it is equivalent to the transfer step, where we do supervised training on the downstream dataset.
}
\label{fig:invalid_supp}
\end{table}

\begin{table}[h!]
\centering
\resizebox{1\textwidth}{!}{
\begin{tabular}{l l c c c c c}
\hline
\textbf{Method}         & \textbf{Data Augmentation}                                                                                                                                                    & \textbf{Epochs} & \textbf{Patience} & \textbf{Batch Size} & \textbf{Optimizer} & \textbf{Learning Rate}       \\
Supervised              & \begin{tabular}[c]{@{}l@{}}Downsample to size 272$\times$224, \\ random crop to size 224$\times$224.\end{tabular}                                                             & 50                  & 5                 & 64                  & Adam               & $[1e^{-5}, \cdots, 1e^{-3}]$ \\
\hline
Unimodal SSL (MoCo-CXR) & \begin{tabular}[c]{@{}l@{}}Random horizontal flip, \\rotation $\leq\pm$10 degrees from \\the original image.\end{tabular}                                                 & 50                  & 5                 & 32                  & SGD                & $[1e^{-5}, \cdots, 1e^{-3}]$ \\
\hline
Multimodal SSL (GLoRIA) & \begin{tabular}[c]{@{}l@{}}Downsample image to size\\ 272$\times$224, random crop to size\\224$\times$224. Extract "impressions"\\section of report for text.\end{tabular} & 50                  & 5                 & 64                  & Adam               & $[1e^{-5}, \cdots, 1e^{-3}]$ \\ \hline
\end{tabular}
}
\caption{
    Implementation details for pretraining (unimodal SSL, multimodal SSL, and supervised) and transfer (supervised, for both LP and LP-FT). Unimodal and multimodal SSL data augmentation, epochs/patience, batch size, and optimizer hyperparameters are taken directly from the respective papers \cite{sowrirajan2021mococxr,huang2021gloria}, and we perform a grid search to determine learning rate (final learning rate chosen using validation loss).
}
\label{fig:implementation_supp}
\end{table}


\begin{table}[h!]
\centering
\resizebox{1\textwidth}{!}{
\begin{tabular}{l | l l l l}
    \hline
    \textbf{\shortstack{Leave-One-Task-Out\\Model Name}} & \textbf{Transfer Dataset} & \textbf{\shortstack{Left-Out\\Pretrain Task(s)}} & \textbf{Transfer Task} \\
    \hline
    "LOTO Pneumonia" & Intermountain & Pneumonia & Pneumonia \\
    "LOTO Pneumonia" & Intermountain & Pneumonia & Pneumonia+Multifocal Pneumonia \\
    "LOTO Pneumonia+Effusion" & Intermountain & Pneumonia,Effusion & Pneumonia+Effusion \\ 
    "LOTO Pneumothorax" & CANDID-PTX & Pneumothorax & Pneumothorax \\
    "LOTO Fracture" & CANDID-PTX & Fracture & Rib Fracture \\
    "LOTO Support Devices" & CANDID-PTX & Support Devices & Intercostal Chest Tube \\ 
    \hline
\end{tabular}
}
\caption{
    To investigate the impact of "novel" tasks for supervised-pretrained models, we create 5 "leave-one-task-out" ("LOTO") models and downstream task pairs described in the above table. For all LOTO models, we pretrain only on CheXpert and use all other CheXpert tasks except the "left-out" ones during supervised pretraining.
}
\label{fig:loto_supp}
\end{table}


\begin{figure}[h!]
    \centering
    \includegraphics[width = 1\textwidth]{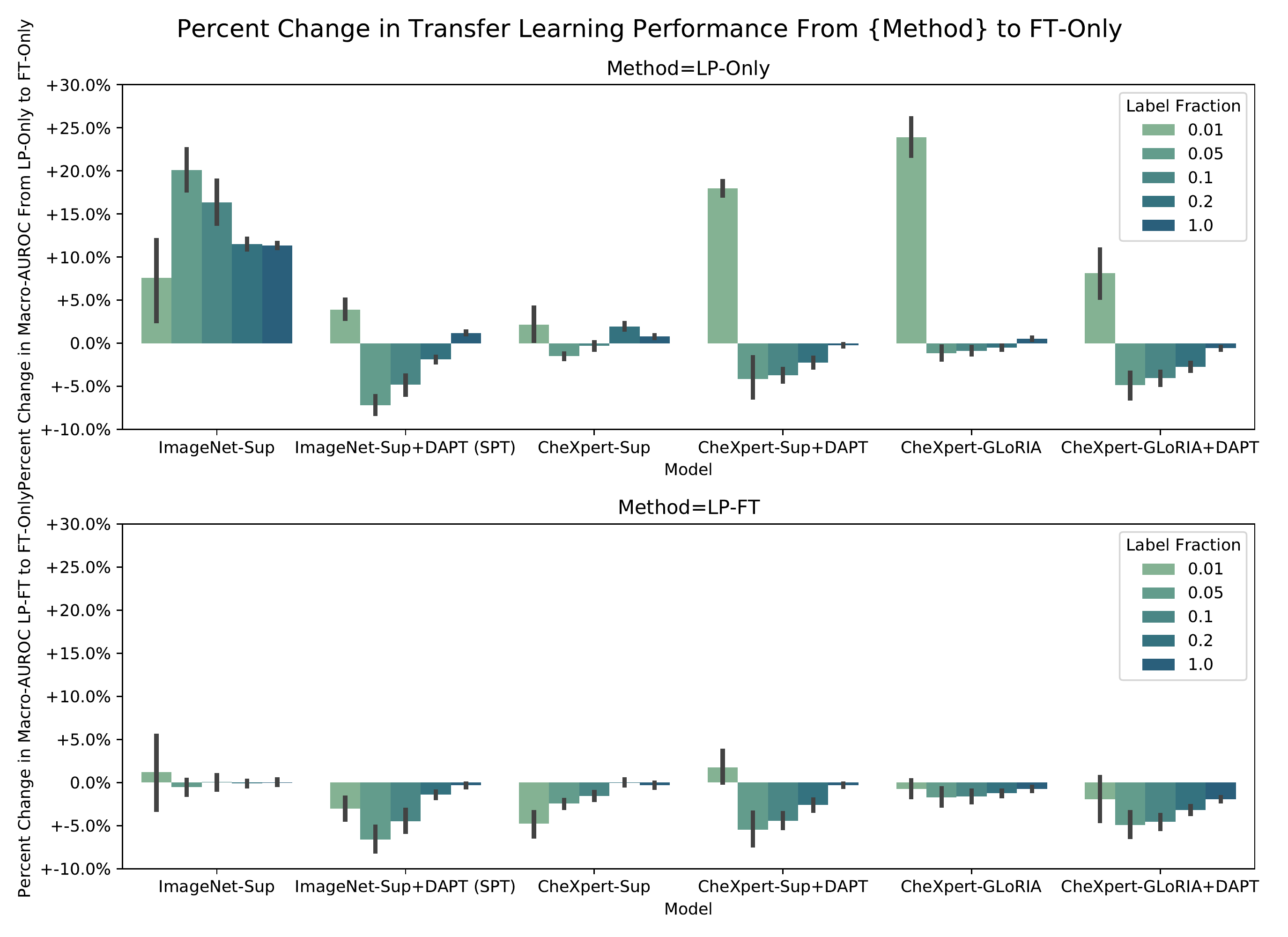}
    \caption{End-to-end finetuning with a randomly-initialized linear classifier (FT) underperforms against LP and LP-FT. Downstream performance difference between LP and FT or LP-FT and FT. Error bars over 5 runs (95\% CI). For LP vs FT, we see increased performance only for lower-performing models: the ImageNet-supervised-pretrained models and models with the lowest transfer learning label fraction (0.01). For more performant models, we see a performance decrease after switching from LP to FT. For LP-FT vs FT, we see matching or decreased performance for all models and label fractions when switching from LP-FT to FT.}
    \label{fig:ft_supp}
\end{figure}